\documentclass[conference]{IEEEtran}
\ifCLASSINFOpdf

\else

\fi

\usepackage[utf8]{inputenc}
\usepackage{graphicx}

\title{Application of Federated Learning in building a robust COVID-19 Chest X-ray classification Model}

\author{\IEEEauthorblockN{Amartya Bhattacharya}
\IEEEauthorblockA{Department of Computer Science and  Engineering\\
University of Calcutta\\
}
\and
\IEEEauthorblockN{Manish Gawali}
\IEEEauthorblockA{Senior Data Scientist\\
DeepTek Inc\\
}
\and
\IEEEauthorblockN{Jitesh Seth\\}
\IEEEauthorblockA{Data Scientist\\
DeepTek Inc\\
}
\and

\IEEEauthorblockN{Viraj Kulkarni\\}
\IEEEauthorblockA{Chief Data Scientist\\
DeepTek Inc\\
}}

\begin{document}

\maketitle

\begin{abstract}
While developing artificial intelligence (AI)-based algorithms to solve problems, the amount of data plays a pivotal role---large amount of data helps the researchers and engineers to develop robust AI algorithms. In the case of building AI-based models for problems related to medical imaging, these data need to be transferred from the medical institutions where they were acquired to the organizations developing the algorithms. This movement of data involves time-consuming formalities like complying with HIPAA, GDPR, etc.There is also a risk of patients' private data getting leaked, compromising their confidentiality. One solution to these problems is using the Federated Learning framework.

 Federated Learning (FL) helps AI models to generalize better and create a robust AI model by using data from different sources having different distributions and data characteristics without moving all the data to a central server. In our paper, we apply the FL framework for training a deep learning model to solve a binary classification problem of predicting the presence or absence of COVID-19. We took three different sources of data and trained individual models on each source. Then we trained an FL model on the complete data and compared all the model performances. We demonstrated that the FL model performs better than the individual models. Moreover, the FL model performed at par with the model trained on all the data combined at a central server. Thus Federated Learning leads to generalized AI models without the cost of data transfer and regulatory overhead.
\end{abstract}

\begin{figure}[htp]
    \centering
    \includegraphics[width=7cm,height = 5 cm]{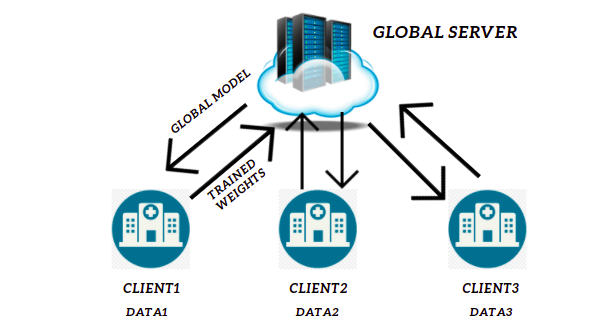}
    \caption{The Federated Learning Process: a) Initially a global model is sent to clients' local servers b) The model gets trained on the local servers c) Model updates are then sent back to the global server d) The global model then utilizes the updates to build a better model and the process continues till we get a robust model}
    \label{fig:galaxy}
\end{figure}

\section{Introduction}
COVID-19  created a huge crisis in the healthcare sector, which was not equipped to handle the situation. With the increase in cases, it became difficult for radiologists to diagnose COVID-19 cases accurately and timely from the thousands of X-Ray reports. These situations suggested the need for automatic detection and classification of X-Ray images.\cite{sheller2020federated} . Thus researchers developed deep learning-based solutions to predict COVID-19 from X-Ray images\cite{zhang2020covid}. 

The creation of these solutions required data to be transferred from medical institutions where it was acquired to organizations working on Computer Aided Detection(CAD) solutions. This process made the data vulnerable to exposure to several third-party individuals or organizations and getting misused.

Moreover, while creating  AI models, generally the dataset used belongs to a single source and hence the data points are sampled from a distribution which might not be representative of the entire population. This could result in poor AI model performance when validating on a dataset with a different distribution from the training set.

Federated Learning (FL)
\cite{konevcny2016federated}\cite{kulkarni2020survey} is a solution to tackle both the problems mentioned above. Federated Learning is a distributed learning framework where models can be trained on the clients' data on their own servers 
without going through the complications of creating the model or distributing their data to any third-party organizations.
In FL the models are trained through multiple Federated rounds which results into a robust model being formed. A single FL round of training works by initially creating a model at a global server. The model is the pushed to all the clients' servers. Since the data is present in the clients' local server, the model gets trained at each of the local server. After training the updated models are then sent back to the global server where the Federated Averaging process\cite{konevcny2016federated} is used to form a better model. In this way a robust model is built which results in better performance irrespective of the source of the test data.

Thus, a robust solution is developed without the involvement of any data transfer.\cite{li2020review}

To this date, the use of FL in the medical domain has been limited. But it has the potential to build AI models which can generalize well and reduce bias
\cite{zhang2021dynamic}\cite{rieke2020future}. In this paper, we applied FL to train a Convolutional Neural Network(CNN) model to classify chest X-rays belonging to COVID-19 affected patients from others.

\section{Related Works}
There is not much work on the application of FL in classification of chest X-Rays (CXRs) into COVID and non-COVID classes. Liu et. al 2020\cite{liu2020experiments}, built a novel model named Covid-Net that performed the task of detecting COVID-19 in CXRs with an accuracy of 92\% with the model having a ResNet18 backbone, which proved to be better than some of the existing state-of-the-art CNN-based models in a federated architecture\cite{feki2021federated}
\cite{zhang2021dynamic}\cite{rieke2020future}. They combined the data taken from different sources and then segregated it into multiple parts to maintain the non-IID characteristics. Feki et.al 2021 \cite{feki2021federated} studied the performance of the Federated Learning model in classifying COVID-19 and non-COVID-19 chest X-Rays.They used a dataset containing 108 COVID-19 cases and then split them equally into 4 clients, but this meant that the data being used was still IID.

We address these shortcomings by taking three different sources of data to maintain the non-IID nature.\cite{gawali2021comparison} The data contains images belonging to two classes i.e COVID-19 present or absent. We trained an AI model on this data using the FLframework. The model is compared with each model trained on a single source and tested on other sources of data. Moreover, we also compared the model built using Federated Learning with a model trained by combining the data from all the sources.

In this paper we have trained five models. On each of the three client servers we trained individual models (Client1, Client2 and Client3). We also pooled all the data to train a centralised global model (the "Combined model"). Then, we used the FL framework to train another global model (the "FL model"). The performances of these models were evaluated on separate test sets from each client. We found that the FL model performed better than all the individual models and at par with the combined model in terms of the AUROC and AUPRC metrics.

\section{Data and Methods}

\subsection{Data}
\begin{figure}[htp]
    \centering
    \includegraphics[width=0.45\textwidth]{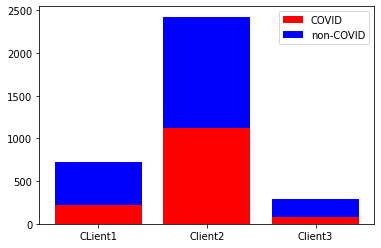}
    \caption{Client-wise distribution of CXRs of patients having or not having COVID-19.}
    \label{fig:distributions}
\end{figure}

\begin{figure}[htp]
    \centering
    \includegraphics[width=0.45\textwidth]{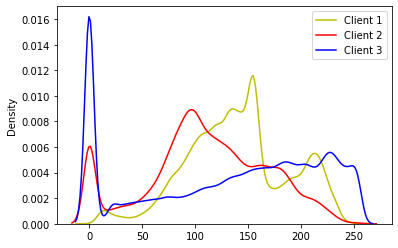}
    \caption{The frequency of pixel values in each dataset. All CXR images had 8-bit encoding and thus values range from 0-255. We clearly see how the data from each client is vastly different from each other.But it should be noted the minimum value of the pixel present was 0, but the curve was extrapolated in order to be converged with the axis. }
    \label{fig:pixeldensity}
\end{figure}

We formulated the experiment by ensuring that the data used for each of the clients is non-Independent Identically Distributed(IID). In our experiments, data from three sources were used. The datasets had variations in the number of images per class i.e. COVID-19 and non-COVID-19. For the Client1, the COVID-19 chest X-ray data was collected from the SIRM website \cite{rahman2021exploring} and the non-COVID-19 chest X-ray data was taken from the RSNA Pneumonia Detection challenge on Kaggle\cite{wang2017chestx}\cite{rahman2021exploring}. The data for Client 1 was downsampled such that it had a total of 719 images out of which 219 belonged to the COVID-19 class and the rest to the non-COVID-19 class. For Client2 the chest X-Ray images were collected from the Eurorad database from Kaggle Repository\cite{chowdhury2020can} . There were 2416 images out of which 1125 belonged to the COVID-19 class. Client3 data was collected from the IEEE data repository\cite{cohen2020covid} having 287 images out of which 84 images belonged to the COVID-19 affected chest X-ray images. The data was then divided into the training, validation, and test datasets for each of the clients. The details about the split of the data for each of the clients are mentioned in TABLE \ref{tab:dist}. A large amount of variation was present in the data in terms of distribution as shown in Fig. \ref{fig:distributions} and the distribution of the pixel values per unit area has been given in \ref{fig:pixeldensity}.

\begin{table}[hbt!]
\begin{center}
\begin{tabular}{|l|l|l|l|l|l|}
\hline
Client   & Data Sources                                                              & Train & Validation & Test & Total \\ \hline
Client 1 & \begin{tabular}[c]{@{}l@{}}Positive - SIRM\\ Negative - RSNA\end{tabular} & 623   & 48         & 48   & 719   \\ \hline
Client 2 & Eurorad                                                                   & 2330  & 48         & 48   & 2426  \\ \hline
Client 3 & IEEE Data Repository                                                      & 191   & 48         & 48   & 287   \\ \hline
Total    &                                                                           & 3144  & 144        & 144  & 3432  \\ \hline

\end{tabular}
\caption{The source and the distribution of the data points (CXRs) after it is split into train, validation and test set }
\label{tab:dist}
\end{center}
\end{table}

\subsection{Methods}
We used DenseNet121\cite{huang2017densely} as the main neural network architecture. At first, initial models were built for each of the clients and trained on their data. Since it was a binary classification problem  i.e non-COVID-19 and COVID-19, the final layer of the neural network model had a single neuron. We used Binary Cross-Entropy as the loss function. We used the Adam optimizer was used with the learning rate fixed to $10^{-3}$ since it was found to converge early. The combined model and the individual models were trained for 20 rounds. We set the number of epochs to 20 as we found out that the model converges within 20 rounds.

The FL model was trained using DenseNet121 as the base architecture. it was initialized using Imagenet weights. The global model was then transferred to be trained on each of the client’s datasets locally. In the training process, the local model was trained for 4 epochs and the weights corresponding to the least validation loss were stored.
These weights were sent to the global server and the global weights were calculated by averaging the local weights. The global FL model was then sent to all the clients for the next FL training round.

This FL training round was repeated 5 times, thus the whole model underwent training for 20 epochs. We used threshold agnostic metrics like ROC-AUC, ROC-PRC scores for comparing the model performances as it is difficult to compare the model performance of different models using threshold dependent metrics like sensitivity, specificity, and F1-score. The results obtained have been discussed below.

\section{Results}
\begin{figure*}[htp]
    \centering
    \includegraphics[width=\textwidth]{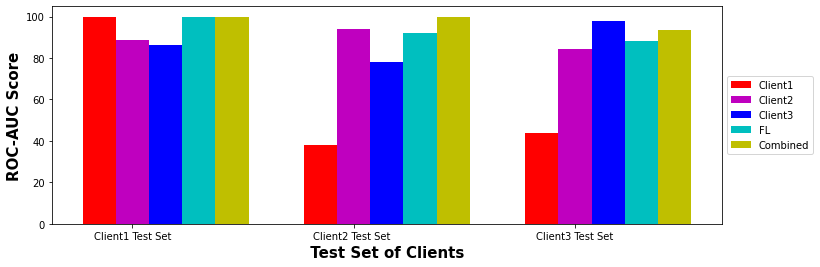}
    \caption{Model Performances using the ROC-AUC scores. In general it can be seen from the plots that FL model performs well in classifying the test data for all the clients than the models that uses only single source of data.}
    \label{fig:auroc}
\end{figure*}

\begin{figure*}[htp]
    \centering
    \includegraphics[width=\textwidth]{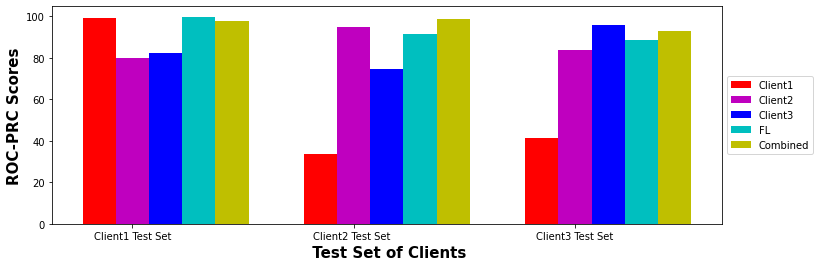}
    \caption{Model Performances using the ROC-PRC scores. In general it can be seen from the plots that FL model performs well in classifying the test data for all the clients than the models that uses only single source of data.}
    \label{fig:auprc}
\end{figure*}
\begin{table}

\begin{tabular}{||c c c c c c||} 
 \hline
 Data  & Client1 & Client2 & Client3 & FL & Combined \\ [0.5ex] 
 \hline\hline
 Client1 & 99.96 & 88.85 & 86.44 & 99.61 & 99.86\\
 \hline
 Client2 & 37.85 & 93.94 & 77.94 &92.31& 99.92 \\
 \hline
 Client3 & 43.85 & 84.57 & 97.73 &88.31 & 93.51\\
 \hline
 
\end{tabular}
\\
\caption{\label{tab:ROC}ROC-AUC Score}

\end{table}

\begin{table}

\begin{tabular}{||c c c c c c||} 
 \hline
 Data  & Client1 & Client2 & Client3 & FL & Combined \\ [0.5ex] 
 \hline\hline
 Client1 & 99.01 & 80.03 & 82.28 &    99.85  & 97.96\\
 \hline
 Client2 & 33.75 & 94.75 & 74.62 &  91.54   &  98.94\\
 \hline
 Client3 & 41.32 & 83.68 & 95.67 &   88.57  & 92.86\\
 \hline
 
\end{tabular}
\\
\caption{\label{tab:PRC}ROC-PRC Score}

\end{table}

The models trained with Client1 data, Client2 data, Client3 data, and the combined model were evaluated against the test data from each of the clients. The results were then compared against the results obtained by the FL Model. Figures \ref{fig:auroc} and \ref{fig:auprc} show the comparison of the ROC-AUC, ROC-PRC scores for each of the models.

From the graph of the performance, it can be observed that each of the clients’ models performed well when evaluated against the test data belonging to the same source as the data with which it was trained. But they failed to perform as well  when they were provided data from a different source. The same observation can be made for both the AUC-PRC table (TABLE \ref{tab:PRC}) as well as the AUC-ROC table (TABLE \ref{tab:ROC}).

Thus it can be concluded  that the models, when trained from a single source of data  get biased and don’t perform well on other sources of data. 

From Fig. \ref{fig:auroc}, Fig. \ref{fig:auprc} and as well as from TABLE \ref{tab:PRC} and TABLE \ref{tab:ROC}, it can be seen that the model trained using the Federated Learning framework solves this issue and performs equally well on the test data coming from different sources. Unlike the individual models, it performs exceptionally well on the test data coming from different sources. Thus it can be inferred that by using the different sources of data and averaging the weights learned during the training process, a more robust model can be achieved.

Moreover, the process of developing a FL Deep Learning model did not require the data from all the sources to be sent for training. Instead, the model was sent to the clients to train the models locally. This suggests that the performance of the Federated Learning model is impressive as compared to the combined model. Although the combined model performs better than the Federated Model, all the data had to be sent to a central location to train it, which compromised the privacy issue as discussed above. 

\section{Conclusion}
AI models tend to get biased towards the training data and don’t perform well when evaluated against the data coming from different sources. This bias can be reduced if we utilize data from different sources for the training of a deep learning neural network model. In the case of medical imaging, the movement of data from medical institutions to organizations developing solutions is a time-consuming process.

Federated Learning is one of the solutions that try to solve both the problems by generalizing the model, by training the model parameters using the data from different sources. In this paper, we applied the Federated Learning-based framework to present a robust solution for classifying COVID and non-COVID chest X-ray images. We trained 5 different models to compare the results. Three of those models were built on the corresponding clients' data, one was built using Federated Learning, and another one by combining all the data. From the results obtained, we could infer that the Federated Learning model performed better than the models built using the three clients' data individually, thus generalizing well. Moreover, performed almost equally well as the model trained on the aggregated data set. This shows that in real-time, Federated Learning shows a path to solve both the privacy as well as generalization issues.

\bibliographystyle{ieeetr}
\bibliography{bibliography}

\end{document}